\pdfoutput=1

\documentclass[11pt]{article}

\usepackage[final]{acl}

\usepackage{times}
\usepackage{latexsym}
\usepackage{paralist}

\usepackage[T1]{fontenc}

\usepackage[utf8]{inputenc}

\usepackage[normalem]{ulem}
\usepackage{microtype}

\usepackage{inconsolata}

\usepackage{graphicx}
\usepackage{multirow}

\usepackage[most]{tcolorbox}

\title{Reading between the Lines: Can LLMs Identify Cross-Cultural Communication Gaps?}

\author{Sougata Saha\textsuperscript{1\thanks{Both authors contributed equally to this paper.}}, Saurabh Kumar Pandey\textsuperscript{1\footnotemark[1]}, Harshit Gupta\textsuperscript{2}, Monojit Choudhury\textsuperscript{1}\\
\textsuperscript{1}Mohamed bin Zayed University of Artificial Intelligence, \textsuperscript{2}IIIT Hyderabad\\
\texttt{\textsuperscript{1}\{sougata.saha, saurabh.pandey, monojit.choudhury\}@mbzuai.ac.ae}\\
\texttt{\textsuperscript{2}\{harshit.g@research.iiit.ac.in\}}
}

\begin{document}
\maketitle
\begin{abstract}
In a rapidly globalizing and digital world, content such as book and product reviews created by people from diverse cultures are read and consumed by others from different corners of the world. In this paper, we investigate the extent and patterns of gaps in understandability of book reviews due to the presence of culturally-specific items and elements that might be alien to users from another culture. Our user-study on 57 book reviews from Goodreads reveal that 83\% of the reviews had at least one culture-specific difficult-to-understand element. We also evaluate the efficacy of GPT-4o in identifying such items, given the cultural background of the reader; the results are mixed, implying a significant scope for improvement. Our datasets are available here: \url{https://github.com/sougata-ub/reading_between_lines}.
\end{abstract}

\section{Introduction}
Whether performing outdoor activities, household chores, or surfing the internet, our perception and understanding of the world are varied. Each of us has a distinct worldview, which causes us to understand and internalize information distinctly. Although our knowledge, and hence our worldview, are individual-specific~\cite{collins1987people, jonassen1999mental, denzau1994shared}, our culture does contribute to the variation by shaping our worldview and understandability~\cite{bender2013cognition, cole2019culture}. This effect of culture is also evident in the online world, where we might find text from distinct online sources hard to understand due to a lack of common ground between us and the writer of the text~\cite{meyer2014culture, korkut2018study}. For example, people unfamiliar with the Arabic culture might not understand the meaning of the dishes "Machboos" and "Luqaimat" from the review text "The Machboos was perfectly spiced, and the Luqaimat was a real treat." Or, someone unfamiliar with the US culture and not exposed to fast food might not understand that "golden arches" refers to "McDonald's" in the text "Let's go to the golden arches for a quick bite.". 

The recent advancements in AI and NLP make us question whether and how Large Language Models (LLMs) can help overcome such barriers in online cross-cultural communication. Here, we consider a use case where LLMs act as cultural mediators by identifying, categorizing, and explaining
text spans \textit{(Culture-Specific Items)} from online reviews that users from distinct cultures might find hard to understand. Although numerous use cases of LLMs in cross-cultural communication are conceivable~\cite{singh-etal-2024-translating, liu-etal-2024-multilingual}, the following fundamental and overarching questions need answering before implementing them across any domain. \underline{Firstly}, although studies discuss the importance of cross-cultural communication \cite{bourges1998meaning}, the need for LLMs and AI assistants in general for cross-cultural communication has not been studied. \underline{Secondly}, the cross-cultural performance parity of LLMs in such settings is unclear and unmeasured. Although studies have shown LLMs to exhibit Anglo-centric biases~\cite{johnson2022ghost, dwivedi-etal-2023-eticor}, this does not mean they perform well in those cultures and poorly in others. We refer to this as the \textit{equitability}, defined as a model's capability to maintain the parity between its performance and the user's need across cultures. Through our specific use-case of LLMs as a cultural mediator, we answer these fundamental questions, a prerequisite of using them to overcome cross-cultural communication barriers. 

The contributions of this paper are as follows:


\begin{compactitem}
    \item We perform a user-study spanning participants from the USA, Mexico, and India to measure their difficulty in understanding Goodreads English reviews of books from the USA, India, and Ethiopia.
    \item We establish the nature and extent of cross-cultural communication gap for Goodreads reviews that AI reading assistants can help bridge.
    \item We evaluate GPT-4o as an \textit{equitable} cultural reading assistant for identifying, categorizing, and explaining culture-specific items from book reviews for urban educated users with different literary preferences from each country.
    \item We share our human (gold) and GPT-4o-annotated (silver) dataset that identifies cultural spans from book review texts\footnote{Dataset available here: \url{https://github.com/sougata-ub/reading_between_lines}}.
\end{compactitem}




\section{Culture: A Primer}
Culture is a complex concept, and is defined as the "Way of life of a collective group of people distinguishing them from other groups" \cite{blake2000defining, monaghan2012cultural, parsons1972culture, munch1992theory}, making it experiential \cite{geertz2017interpretation, Bourdieu_1977}. It provides a common ground to groups through shared experiences and creates distinct "worldviews" between people from different cultures \cite{collins1987people, jonassen1999mental, denzau1994shared, bang2007cultural, mchugh2008cultural} by shaping their cognition \cite{mishra2001cognition, nisbett2002culture, bender2013cognition, cole2019culture}. 
This difference in perception due to culture affects all aspects of life, where literary works composed by individuals from one culture might not be perceived similarly by individuals from a different culture and hamper cross-cultural communication \cite{meyer2014culture, korkut2018study}. In NLP, these variations can broadly be due to differences in linguistic form and style, common ground, aboutness, and values \cite{hershcovich-etal-2022-challenges}. Among demographies, these differences vary across dimensions or \textit{semantic proxies}, ranging from physical items such as food and materials to abstract constructs like emotions and values \cite{thompson2020cultural, adilazuarda-etal-2024-towards}. \citet{newmark2003textbook} terms such concepts as \textit{Culture-Specific Items (CSIs)}, identifying and addressing which is crucial for cross-cultural communication.
Thus, to technologically facilitate cross-cultural communication, it is essential to identify and gauge the extent of the variation of such CSIs, which can inform appropriate technological advancements for mitigation.

\section{User Study}

To measure the extent of the difficulty in comprehension due to culture and establish a need for cross-cultural tools, we conducted an online user study where we showed participants book reviews written in English and asked them to highlight text spans they did not understand. The length of the spans was unrestricted and could be individual words or even multiple sentences. Our goal was to quantify how much of things people do not understand and, out of that, how much is cultural (CSIs). Measuring this cultural knowledge gap can serve as an evaluation benchmark for tools that facilitate cross-cultural communication. 

\subsection{Dataset}

The study was performed over 57 English review texts of books originating from Ethiopia, India, and the USA, which represent distinct cultures per Inglhart-Welzel's world cultural map \cite{inglehart2010wvs}. The reviews were sampled from the Goodreads dataset collected by \citet{goodreads_data_1, goodreads_data_2} in 2017, a sizeable publicly available book-review dataset spanning 2.3 million books from multiple countries and languages. We sampled the 57 reviews in two steps. First, we prompted GPT-4o \cite{achiam2023gpt} to list the top 20 famous authors of fictional and non-fictional books from each country and restricted to reviews of books from these authors with at least 50 reviews and review lengths between 50 and 200 words. We randomly sampled 3,100 reviews while ensuring at least 250 reviews from each country. Next, to ensure enough cultural diversity in our study, for each review text, we prompted GPT 3.5 to identify CSIs for people from our country of study, India, Mexico, and the USA, with varied literary preferences: fiction or non-fiction. Section \ref{prompt_engineering} discusses the prompt in detail. The model identified CSIs in almost all review texts, with varied levels of familiarity. We randomly sampled 57 reviews containing at least one unfamiliar CSI as the final dataset for our study. Table \ref{tab:data-stage-stat} shares the data statistics after each filtering step.




\begin{table}[h]
\centering
\resizebox{\columnwidth}{!}{%
\begin{tabular}{|l|l|l|l|c|}
\hline
\textbf{Country} & \textbf{Genre} & \textbf{\# Reviews} & \textbf{Avg Length} & \multicolumn{1}{l|}{\textbf{Avg \# CSI}} \\ \hline
Ethiopia & fiction     & \ \ 206 | 6   & 100.8 | 73.7  & 6.2 | 5.0 \\
Ethiopia & non-fiction & \ \ \ \ 51 | 8    & 102.3 | 101.4 & 6.2 | 6.8 \\ \hline
India    & fiction     & \ \ 393 | 7   & 101.9 | 98.6  & 5.9 | 5.6 \\
India    & non-fiction & \ \ 209 | 11  & 100.9 | 96.4  & 6.6 | 6.1 \\ \hline
USA      & fiction     & 1487 | 15 & 104.5 | 91.1  & 6.0 | 5.8 \\
USA      & non-fiction & \ \ 752 | 10  & 101.8 | 134.6 & 6.3 | 5.8 \\ \hline
\end{tabular}%
}
\caption{Review statistics after each filtering step (1 | 2)}
\label{tab:data-stage-stat}
\end{table}

It is worth mentioning that from the perspective of generalizability, the current sample size suffices because we did not seek to establish concrete quantitative bounds on cross-cultural readability. Also, since there are no other studies of such kind, it is hard to determine an upper limit on the scale of the study while also factoring in the cost. Our objectives are primarily to motivate the need for cross-cultural AI-based reading assistants by estimating how much people do not understand due to culture and benchmarking how LLMs such as GPT-4o would perform as cultural AI assistants.

\subsection{Method}
\subsubsection{Pilot Studies}
\label{pilots}
We conducted two small-scale internal and one external pilot study to determine the questionnaire\footnote{The experiments are approved by MBZUAI's internal IRB, bearing case number 7.}.

\noindent
\textbf{Pilot 1: } Similar to the task formulation for GPT 3.5 in the Step 2 filtering, in the first pilot, we showed a review text to six participants and asked them to identify spans mentioning CSIs that they did not understand or were unfamiliar with, given their cultural background. The participants were researchers from India, Ethiopia, and the USA. The key takeaways from the study were: (i) Although the participants cumulatively identified 13 CSIs (detailed in Table \ref{tab:pilot_1_csis} Appendix \ref{sec:appendix}), the question forced ethnocentrism, where one had to internally assume a source culture of the review text before determining CSIs that might be common in the source culture but exotic to their culture. (ii) Such a formulation forced participants to generalize their culture, which they found difficult. (iii) It is difficult to distinguish the unknowability of CSIs borne out of cultural difficulty and otherwise.

\noindent
\textbf{Pilot 2: } Incorporating their feedback, we reformulated the task in Pilot 2. Instead of asking what one does not understand due to their culture, we asked participants to identify spans they generally found hard to understand or were unfamiliar with. Additionally, we asked several other questions (detailed in Appendix A) with multiple choices to capture their understanding of the review from different aspects and provoke their thoughts around the task. Pilot 2 encompassed three internal participants, and each reviewed five texts. Overall, we observed from the study that (i) The participants found the task more natural than before as the formulation did not enforce taking any cultural viewpoint. (ii) Since the questions revolved around general understandability, they highlighted spelling mistakes, complex words, and vocabulary as things they did not understand. (iii) As depicted in Table \ref{tab:pilot-2-csis}, the length of the spans highlighting the same concept varied by participant.

\noindent
\textbf{Pilot 3: } Incorporating the feedback from Pilot 2 participants, we refined a few questions and conducted a pilot study in Prolific with 13 participants from India and the USA. We used Google Forms for the survey, comprising a single review text. We observed that (i) Similar to Pilot 2, the highlighted span length varied between annotators for the same concept, where some annotators marked the entire review text instead of only highlighting spans. (ii) The annotations varied greatly. Some annotators did not find any concepts difficult, and some marked the entire text as hard. A few participants highlighted the author's name and book title in the review text as difficult-to-comprehend concepts.

Since performing span-level annotations is difficult in Google Forms, we transitioned to Label Studio as the annotation interface to ease the annotation difficulty in the actual experiments.

\subsubsection{Actual Study}
We used Label Studio \cite{Label} with Prolific\footnote{\url{www.prolific.com}} for the actual study. Our goal was to collect annotated text spans from review texts that the participants found difficult to understand or were unfamiliar with. In addition to this, we asked follow-up questions about the review texts to gather further insights. We hosted multiple instances of Label Studio, each secured with individual login credentials, and provided them to annotators through Prolific. Once participants logged into their assigned instance, they were directed to the annotation interface, as shown in Appendix \ref{fig:over_overview}.

The study first explicitly sought consent from the participants, allowing only those who agreed to proceed. Following this, they completed screener validation questions designed to assess their backgrounds, including where they spent most of their time before the age of 18 and their highest level of education. Next, the study asked demographic and preference questions like their age, reading habits, and literary genre preferences. The consent forms, screener validation questions, and demographic and preference forms are in Appendix \ref{fig:1_c_sv_di}. The survey questions were carefully constructed by adhering to established methodologies and best practices by \citet{article} that ensure comprehensive coverage of relevant variables and a high-quality data annotation.

Next, participants were shown a review text and asked if they had read the reviewed book and if they were familiar with the author and their other works. Next, the study asked them to mark the spans of text they did not understand, using different levels of unfamiliarity. Additionally, the study asked five more questions related to the review text to capture their understanding of the review text from diverse aspects. The span-marking task and the seven questions about the review text are in Appendix \ref{fig:span_mark_other_ques}. After completing the annotations, the study directed participants to the end of the survey section (depicted in Appendix \ref{fig:eos_interface}), which shared a survey code to collect their reward in Prolific. Participants were also provided with a detailed guideline document to assist them in performing the annotations, ensuring they followed all necessary principles following Prolific's and our requirements.

The study was completed by 50 participants: 8 from India, 22 from Mexico, and 20 from the USA. To prevent fatigue, which could reduce the quality of responses, we split the data annotation tasks into manageable batches of 12 data points per batch. Participants were paid 7.5 to 8 GBP per batch, and the median time to complete a batch of annotations was approximately 30 to 40 minutes. This payment structure aligns with Prolific's and country-wise standard wage guidelines, ensuring fair compensation for their time and effort. We also ensured that each data point received annotations from multiple annotators, ensuring the reliability and diversity of the collected data. Appendix \ref{appendix:demographics_usa_mex_ind} shares additional demographic details of the participants.


\begin{figure*}[!t]
    \centering 
    \includegraphics[width=\linewidth]{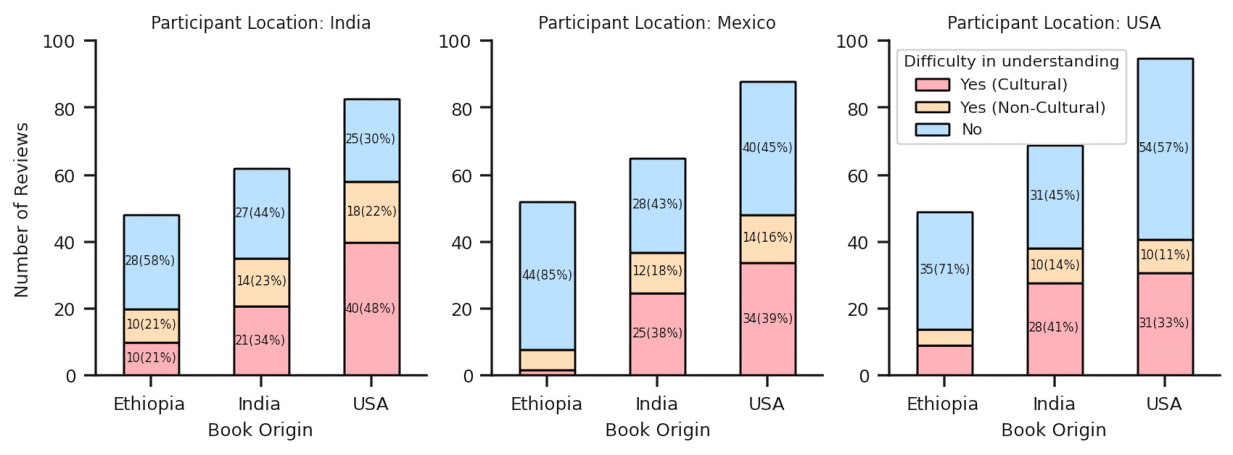} 
    \caption{Participant location-wise difficulty in understanding book reviews by country of origin of the book.}
    \label{fig:understandibility_splits}
\end{figure*}

\subsection{Post Processing}

We collect a total of 761 responses across all three countries. Although crowdsourcing platforms make it easy to conduct user studies at scale for specific demography, the quality of the annotations is sometimes questionable \cite{hsueh2009data, huynh2021surveynlprelatedcrowdsourcinghits, eyal2021data, douglas2023data}. We perform the following quality checks and filter out bad-quality answers by removing responses where (i) People mention they do not understand or are not familiar with certain aspects of the review text but do not mark any spans. (ii) People mention they understand everything and are familiar with everything but still highlight spans. The filtration step removes 11\% of cases, yielding 668 responses.


\subsubsection{LLM-assisted post processing}
Since the study only asked participants to identify spans they found hard to understand, we processed the results using GPT-4o to categorize them as CSIs (cultural spans) or non-CSIs. We first created a taxonomy by interactively prompting ChatGPT using the template outlined in Appendix \ref{fig:init_taxonomy}. We first provided a few review texts and their highlighted spans and instructed ChatGPT to generate categories and subcategories as the initial taxonomy. Then, we iteratively fed the model its proposed taxonomy along with marked spans from an additional 20 annotators, enabling it to continuously update and refine the taxonomy based on this new input. This iterative process allowed the model to enhance the taxonomy effectively.

The final taxonomy classified each span into "Cultural and Linguistic," "Cultural and Non-Linguistic," "Non-Cultural and Linguistic," "Non-Cultural and Non-Linguistic," and "Poor Annotation", listed in Appendix \ref{fig:final_span_categorization}. Adhering to the human-in-the-loop validation process by \citet{shah2023using}, we consulted two linguistic experts to review and validate the generated taxonomy. This step ensured clarity and robustness, preventing over-generalization and ambiguity in the definitions and confirming that our taxonomy effectively captures the diversity and complexity of the spans while maintaining consistency across annotations.

We used GPT-4o to annotate each identified span using our taxonomy. The model inputs each review text and a list of all spans highlighted as non-understandable by participants from any country and associates each span with one of the five classes. Also, since the pilot studies indicated that human annotations tend to be noisy, with some marked spans being superfluous while some reflecting low-quality annotations, we instructed the model to cluster the spans semantically based on the context. This clustering process helped filter out low-quality and irrelevant annotations while resolving ambiguities related to span lengths, allowing us to focus on meaningful data. Appendix \ref{fig:final_span_categorization} depicts the prompt used to post-process using GPT-4o.

\subsubsection{Evaluation}
\label{gpt_error}
The GPT-4o annotations were evaluated by two human experts comprising graduate students from Linguistics and Computer Science. Each expert evaluated 120 random annotations and had 54 overlapping samples. They assessed the span type and the cluster assigned by GPT-4o by marking the correctness with binary flags. The evaluators agreed on 48 out of 54 (89\%) cases for span type and 52 out of 54 (96\%) for the cluster assignment. For the 54 overlapping annotations, at least one evaluator found the span type labels correct in 94\% of cases and the cluster labels correct in all cases.  Although there is an approximate 10\% margin of error for incorrect span type classification and subsequently 15\% error margin for cluster assignment, the high agreement scores indicate that GPT-4o's annotations are reliable. Hence, we cluster all spans identified by the participants from the study as per the clusters assigned by the model. Henceforth, we perform all span-level analysis on its cluster representative.
\begin{figure*}[!t]
    \centering 
    \includegraphics[width=\linewidth]{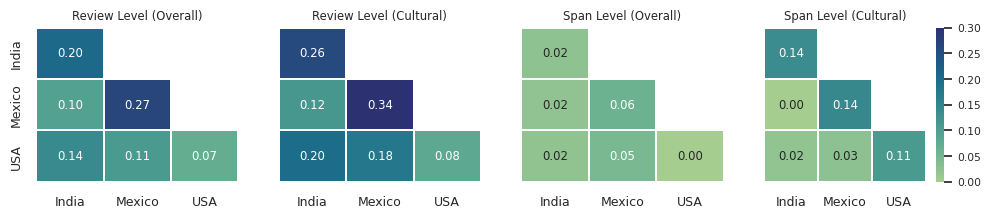} 
    \caption{Inter Annotator Agreement at Review Level and Span Level across Countries.}
    \label{fig:iaa}
\end{figure*}

\subsection{Analysis}
\textbf{1. What is the overall and cultural difficulty of understanding concepts from reviews?}
Out of 611 total evaluations encompassing all participants from all countries, 312 (51\%) evaluations had no difficult-to-understand spans highlighted. The remaining ones had at least one difficult-to-understand span, and 200 (33\%) had some culturally difficult spans. At a review level, all 60 reviews had at least one difficult-to-understand element, where 50 reviews (83\%) had culturally difficult-to-understand elements, while others were non-cultural. On average, a participant evaluated 12 reviews, of which six were hard to understand. Four of the six hard-to-understand reviews had cultural elements.

\noindent
\textbf{2. What is the overall difficulty by each culture and how much of it is cultural?}
Figure \ref{fig:understandibility_splits} plots the proportions of hard-to-understand evaluations by a participant's location and the book's country of origin. We observe the following: (i) Participants from all locations found the reviews from Ethiopia easier to understand and did not contain difficult spans, and evaluators from Mexico did not find such spans in 85\% of cases. Also, in each country, the proportion of the evaluations highlighting culturally difficult spans was lower than the reviews of books from other countries. (ii) Considering participants and books from India and the USA, culturally difficult spans are lowest for the participants from their own country. For evaluators from India, 34\% of unfamiliar spans from the reviews of Indian books were cultural, compared to 48\% for USA book reviews. For the USA, 41\% of unfamiliar spans from the reviews of Indian books were cultural, compared to 33\% for USA book reviews, (iii) For evaluators from Mexico, the proportions of evaluations comprising non-difficult spans and culturally and non-culturally difficult spans were similar for books originating from India and the USA. (iv) Overall, the highest proportion of evaluations highlighting culturally difficult spans were for US-origin books by evaluators from India (48\%), which was followed by India-origin book reviews by evaluators from the USA (41\%).

This cross-cultural gap in understanding indicates that people from all locations might benefit from tools such as cultural reading assistants and translation systems that enable communication across cultures. Also, the cultural difficulty in understanding book reviews from one's own country indicates the dynamic nature and long tail of culture, where culture varies within a country, which also motivates the need for such tools.

\begin{figure*}[!t]
    \centering 
    \includegraphics[width=\linewidth]{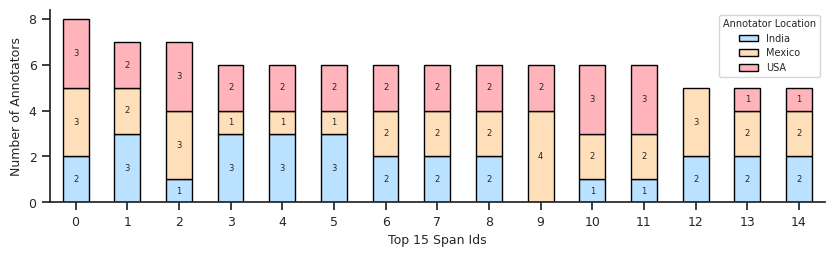} 
    \caption{Top 15 Span Clusters identified by participants across countries. Corresponding span text in Table \ref{tab:top_15_spans}.}
    \label{fig:top_15_spans}
\end{figure*}

\noindent
\textbf{3. How much of the reviews and spans are cultural?}
We measure the review and span-level agreements between participants from the same country and compare them against the agreements when the participants are from different countries. We create all combinations of inter and intra-country annotators and compute Krippendorf's alpha to measure the inter-annotator agreement within a country and between each pair of countries. Figure \ref{fig:iaa} captures the agreement between annotators within a country and across countries. We observe the following at a review level: (i) Annotators from the same country agree more amongst themselves on which reviews are generally hard to understand than annotators across countries except for the USA. We observe a similar trend for reviews containing CSIs, where within a country, annotators agree more on which reviews contain difficult CSIs than inter-country. (ii) For generally difficult and CSI-containing reviews, participants from Mexico agree more among themselves (0.27/0.34) than with reviewers from India and the USA. (iii) Similarly, Indian participants agree more amongst themselves (0.2/0.26) than from other countries. However, they align more with the USA (0.14/0.2) reviewers than Mexicans (0.10/0.12). (iv) Interestingly, reviewers from the USA agree more with Indian and Mexcian participants than among themselves. (v) At a span level, there is a lack of consensus across all countries, which testifies that understandability is individual-specific, which was already indicated by our pilot studies. Also, the confounding factors for the low value are likely due to errors from the GPT-4o post-processing, as mentioned in Section \ref{gpt_error}. However, annotators from the same country seem to agree more on culturally difficult spans than other kinds of spans. The intra-country agreements at the "overall" and "cultural" levels significantly increased from 0.02 to 0.14 for India, 0.06 to 0.14 for Mexico, and 0 to 0.11 for the USA, indicating that although different annotators might find different things hard to understand, there is a level of consensus on CSIs that they find hard to understand. This also implies that CSIs are a set of harder-to-understand constructs in text and, therefore, are good targets for circumventing the cold-start problem \cite{hu2008collaborative} of personalization for user-facing systems. Personalization systems can use such culture-specific information when it does not know anything about a user. Such systems can use culture-specific information when user-specific details are unavailable.

\noindent
\textbf{4. What are the most common spans identified by people from each culture?} 
Figure \ref{fig:top_15_spans} plots the top 15 spans frequently highlighted by participants as hard to understand. Except for a few cases, participants from all countries find these spans hard to understand. Table \ref{tab:top_15_spans} lists the corresponding span texts of the figure. Table \ref{tab:top_csi_countrywise} lists the most frequent unfamiliar text spans by each participant location. Interestingly, cultural spans referring to "Al Gore" and "Hemmingway" are frequently marked as unfamiliar by participants from India, whereas participants from the USA and Mexico find concepts like "topper from engineering institute," which are colloquial terms frequently used in India, as hard to understand.
Given that the understandability of text varies by region and culture, we next experiment with GPT-4o and benchmark its performance as a cultural assistant.

\section{Benchmarking GPT-4o as a Cultural Assistant}
We designed a prompt to evaluate GPT-4o \cite{achiam2023gpt} as a cross-cultural reading assistant by assigning it the role of a cultural mediator which understands all cultures in the world. We tasked the model to identify, categorize, adapt, and explain CSIs from book reviews from unknown source cultures to the target culture of the reader defined by their country and book-genre preference.


\subsection{Experiment Setup}
\label{prompt_engineering}
We carefully constructed a prompt encompassing the desired attributes of a cross-cultural assistant. As depicted in Appendix~\ref{fig:cultural_mediator_prompt}, we task GPT-4o with the following tasks: (i) Identifying and explaining all CSIs from the review text that may be difficult to understand due to a user's cultural background. (ii) For each CSI, identify its category from one of the five categories by \citet{newmark2003textbook}. Following \citet{singh-etal-2024-translating}, we additionally added two extra categories: linguistic and others. (iii) For each CSI, identify its familiarity as one of the following four levels: (a) Familiar- Known by most people, (b) Somewhat familiar- Known by some people, (c) Unfamiliar- Unknown to most people, and (d) Ambiguous- Known but in an opposite way. (iv) Identify the impact on the comprehension of the main review point and reformulate the entire text per the user's background. Since prompting LLMs to generate explanations of their response has shown success~\cite{wang2022self, wang-etal-2024-boosting-language}, we additionally prompt the model to reason their identification of each CSI.

We use the Azure OpenAI platform for our experiments and set the model's temperature to 0. We prompt GPT-4o to obtain outputs in the required format as detailed in Figure~\ref{fig:cultural_mediator_prompt}. We perform post-processing on the obtained response. We remove any JSON or Python markers in the response and convert the response to a dictionary with the required keys. If the post-processing does not yield the response in the required format, we again prompt GPT-4o to format the response.

\begin{figure}[h]
    \centering 
    \includegraphics[width=0.75\linewidth]{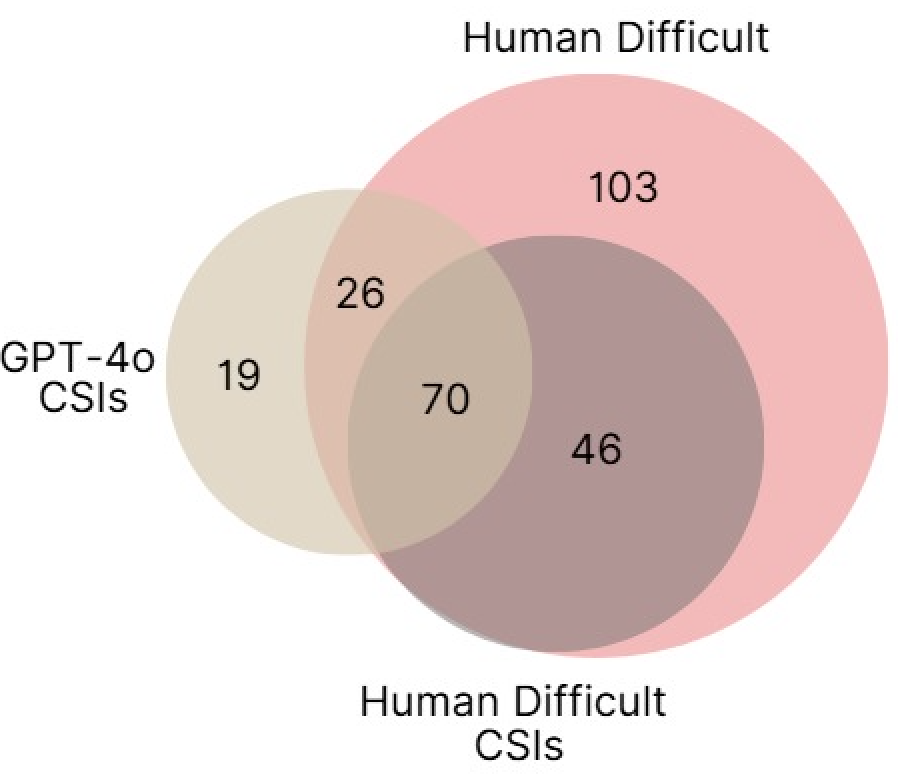} 
    \caption{Overlap between Human-identified difficult spans and GPT-4o-identified CSIs}
    \label{fig:gpt4o-venn}
\end{figure}

\subsection{Analysis}
\begin{figure}[!h]
    \centering 
    \includegraphics[width=0.9\linewidth]{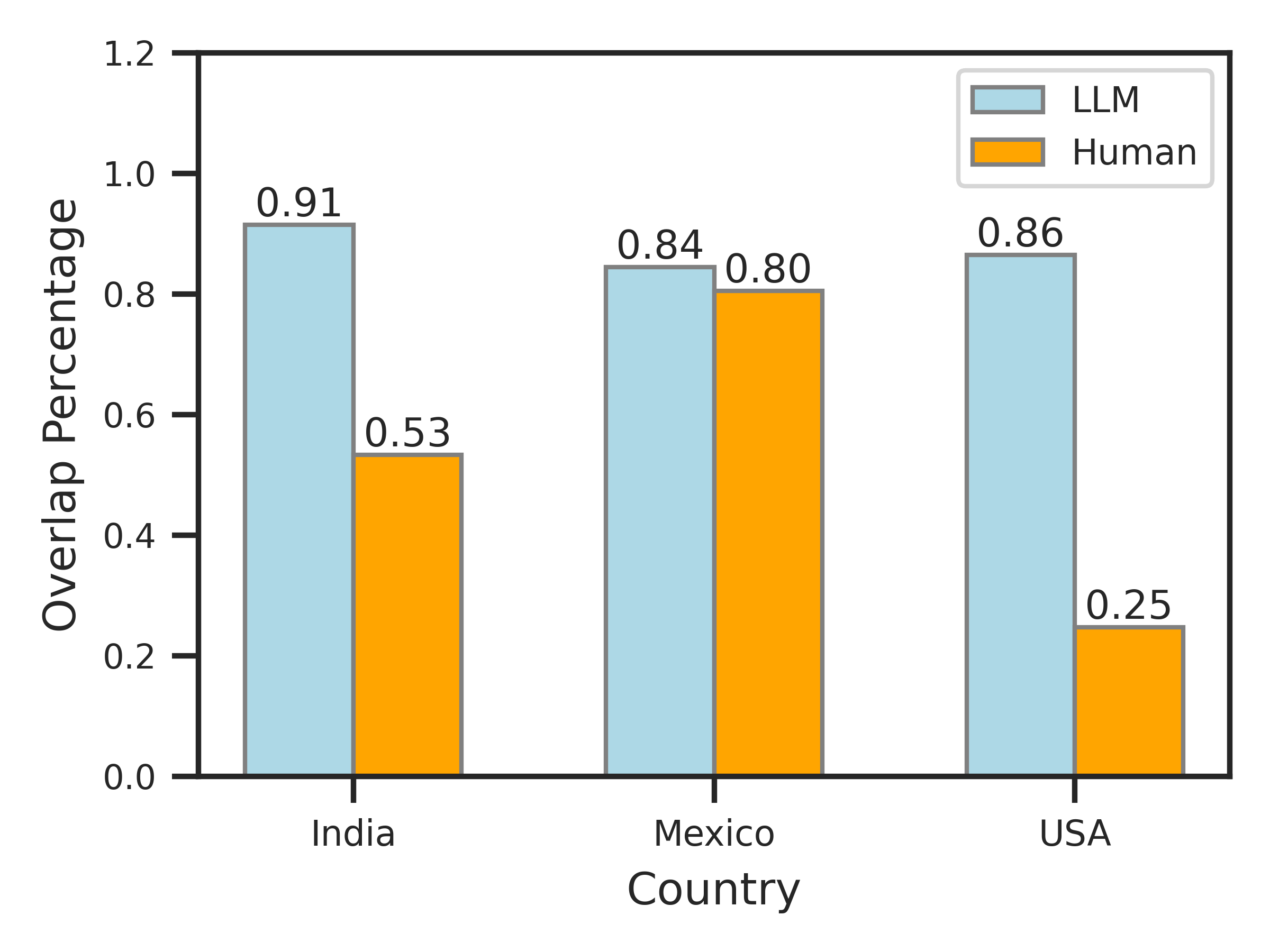} 
    \caption{Overlap percentage of fiction and non-fiction spans across countries.}
    \label{fig:fict_nf_overlap}
\end{figure}

We cluster the GPT-4o-identified spans by matching their embedding-based cosine similarity scores \cite{reimers-2019-sentence-bert} with the span clusters from the user study (similarity threshold=0.5 and measured their overlap. As depicted in Figure 1, combining all the CSIs identified by GPT-4o across all countries and genre preferences, 96 (83\%) overlap with the 245 unique spans marked as difficult to understand by humans, and 70 (60\%) overlap with the 116 user-identified culturally difficult spans. Interestingly, 26 (22\%) of the GPT-4o-identified CSIs, although considered difficult, are not deemed cultural by humans. We hypothesize this to be confounded by two things. First, as discussed in Section 4, the GPT-4o-based cultural classification of the user-identified spans has an error margin of approximately 15\%. Second, since the user study encompassed 50 participants, their identified spans most likely do not capture all the difficult spans for someone from the same culture, contrary to the CSIs identified by GPT-4o, which are more encompassing. This generalization of the model is further evident from Figure \ref{fig:fict_nf_overlap}, which measures the overlap of spans between fiction and non-fiction readers from each country. For all countries, the overlap between GPT-4o-identified spans for both groups is consistently higher than those identified by the users. Table \ref{tab:human_noncultural_llm_cultural_filtered} lists a few such spans.

\begin{figure*}[!t]
    \centering 
    \includegraphics[width=\linewidth]{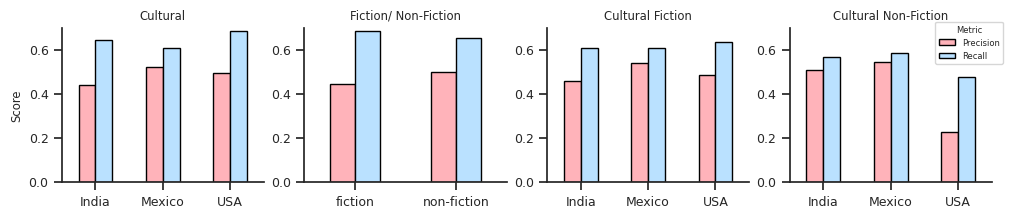} 
    \caption{Precision and recall of the overlap between user-identified culturally difficult spans and GPT-4o-identified CSIs}
    \label{fig:gpt4o-pr-rl }
\end{figure*}

Considering the user-identified cultural spans as the ground truth, we compute the precision and recall scores of GPT-4o-identified CSIs at different levels of country and genre-preference combinations. Figure \ref{fig:gpt4o-pr-rl } plots the scores, where we observe that (i) For all countries, GPT-4o's recall (approx. 0.65) is higher than the precision (approx. 0.49), signifying that the model is generally able to capture a variety of CSIs. Also, the recall for all countries is comparable. (ii) We see a similar trend across genre preferences and combinations of country and genre, where the recall values are higher than precision, signifying the model's variational awareness. However, the model's performance drops for non-fiction readers from the USA. 

Our results indicate that GPT-4o seems equitably aligned with users from all countries. However, with an overall precision of 0.49 and recall of 0.65 the magnitude of the alignment is low, indicating much room for improvement.



\section{Discussion}
Although our study encompasses Goodreads book reviews, we believe the findings apply to other domains. Since the books and their reviews were random, they do not pertain to any single domain and often comment on different aspects of life. Our studies suggest that although understandability varies by person, CSIs are an agreed set of items that people do not understand within a culture. Furthermore, the CSIs are distinct across cultures. This finding can benefit any domain and application that spans cultures, such as translation, education, intercultural training, content personalization, etc. In a rapidly globalizing world, where the definition of culture has shifted from a traditional nationality or ethnicity-centric definition to include digital communities \cite{birukou2013formal}, our study establishes the need for tools that enable cross-cultural communication and demonstrates where current state-of-the-art LLMs, such as GPT-4o, are lacking. Recent tools such as Culturally Yours \cite{pandey-etal-2025-culturally}, an AI-based reading assistant that enables cross-cultural communication and personalizes to each user, further attest to the practical implications of our study.

\section{Related Work}
Communicating across cultures is crucial in today's global world. Although several theories and frameworks exist that define the nature of cross-cultural communication \cite{gudykunst2003cross, tannen1983cross, hurn2013cross, gardner1962cross}, practical applications implementing such concepts are still nascent. Computationally, a considerable amount of work has been done in HCI in defining the considerations of cross-cultural tools and systems \cite{bourges1998meaning, kyriakoullis2016culture, heimgartner2018culturally}. Cultural adaptation is also extensively studied in machine translation \cite{bassnett2007culture, trivedi2007translating, sperber1994cross}, \citet{aixela1996culture} introduced the term "Culture-Specific Items" (CSIs), and \citet{newmark2003textbook} further presented a taxonomy of CSIs and proposed methods to tackle each type of CSI. Recently, \citet{zhang-etal-2024-cultural} introduces the ChineseMenuCSI dataset where they integrate the translation theories to create an annotated dataset with CSI vs Non-CSI labels. \citet{singh-etal-2024-translating} leverages open-sourced LLMS to adapt CSIs from the USA TV series Friends, for an Indian audience. 

Advances in LLMs have garnered studies that detect biases in LLMs \cite{tao2023auditing, kharchenko2024llmsrepresentvaluescultures,duan2023ranking, lin-etal-2025-investigating}. Using curated cultural datasets, most methods probe LLMs and test their knowledge and reasoning capabilities in culture-specific settings  \cite{nadeem-etal-2021-stereoset,nangia-etal-2020-crows, wan-etal-2023-personalized, jha-etal-2023-seegull, li2024culturegenrevealingglobalcultural, cao-etal-2023-assessing, tanmay2023probingmoraldevelopmentlarge, rao-etal-2023-ethical, kovač2023largelanguagemodelssuperpositions}. Some methods \cite{kharchenko2024llmsrepresentvaluescultures, li2024culturellm, dawson2024evaluatingculturalawarenessllms} also analyze the model-generated responses along theoretical frameworks such as Hofstede's cultural dimensions \cite{book1, book2} and measure their proximity with cultures, where high proximity indicates better value alignment between the nearby cultures and the values portrayed by the model's response. Most of these methods necessitate constructing cultural-specific test beds \cite{wang-etal-2024-cdeval, rao2024normadbenchmarkmeasuringcultural, NEURIPS2024_8eb88844, zhou2024doesmapotofucontain, putri-etal-2024-llm, mostafazadeh-davani-etal-2024-d3code,  wibowo-etal-2024-copal, owen2024komodolinguisticexpeditionindonesias, chiu2024culturalbenchrobustdiversechallenging, liu-etal-2024-multilingual, koto-etal-2024-indoculture}.

\section{Conclusion}

Although cultural knowledge is necessary for LLMs to operate in cross-cultural settings, its impact on practical cross-cultural settings is unmeasured. Hence, determining LLMs' cross-cultural equitability primarily based on their factual knowledge about cultures is limiting. To address this gap, here we take a distinct approach where we perform a thorough user study of the cross-cultural understandability of Goodreads reviews to understand the need for cross-cultural AI tools. We then evaluate GPT-4o as a cultural reading assistant that can identify and adapt culture-specific items to the culture of a user. Our evaluations reveal that there are indeed lots of cultural things in Goodreads reviews that might not be comprehensible to people from distinct cultures. Also, GPT-4o performs equitably across the studied cultures. However, with overall low precision and recall scores, the amount of alignment is low, indicating room for improvement.


\section*{Limitations}
In this work, we perform a user study on 57 review texts from Goodreads. We limit our study to demographics of India, USA, and Mexico, in the English language. We consider genre preference as the only semantic proxy under study. A large-scale multilingual study across more diverse cultures worldwide and at a larger intersection of demographic and semantic proxies would help strengthen the findings of our work. We limit our evaluation to using GPT-4o since it is known to be better than other models, a full-fledged assessment of other closed-source and open-source models needs to be performed.

\section*{Ethical Considerations}
We do not capture any personally identifiable information regarding the users involved in user study. We follow standard ethical guidelines~\cite{rivers2014ethical} and do not attempt to track users across sites or deanonymize them. We also adhere to the minimum wage policies of all demographics involved in the user study. We filter out all the offensive/adult content from the review texts presented to the users. We plan to release our data and the annotations to foster further research on developing AI assistants for efficient cross-cultural communication.

\section*{Acknowledgements}
This research was supported by the Microsoft
Accelerate Foundation Models Research (AFMR) Grant. We thank all the team members involved in the internal pilot studies. 

\bibliography{acl_latex}

\clearpage

\appendix

\section{Appendix}
\label{sec:appendix}

\begin{table}[h!]
\centering
\resizebox{\columnwidth}{!}{%
\begin{tabular}{|l|c|c|c|c|c|c|}
\hline
\textbf{Identified CSI} & \textbf{Customs} & \textbf{Habits} & \textbf{Linguistic} & \textbf{Material} & \textbf{other} & \textbf{Social} \\ \hline
\begin{tabular}[c]{@{}l@{}}crazy damsel \\ in distress\end{tabular}        &   &   & 4 &   &   &   \\
kickass model                                                              &   &   & 4 &   &   &   \\
\begin{tabular}[c]{@{}l@{}}purple for my \\ taste\end{tabular}             &   &   & 4 &   &   &   \\
gateway book                                                               &   &   & 2 &   &   &   \\
\begin{tabular}[c]{@{}l@{}}paranormal/\\ vampire\\ element\end{tabular}    & 1 &   & 2 &   & 2 &   \\
Harlequins                                                                 &   &   & 1 & 3 &   &   \\
\begin{tabular}[c]{@{}l@{}}leave much \\ room\end{tabular}                 &   &   & 1 &   &   &   \\
Minus                                                                      &   &   & 1 &   &   &   \\
\begin{tabular}[c]{@{}l@{}}romance for \\ the ages\end{tabular}            &   &   & 1 &   &   &   \\
\begin{tabular}[c]{@{}l@{}}slow and \\ atmospheric\end{tabular}            &   & 1 & 1 &   &   &   \\
\begin{tabular}[c]{@{}l@{}}weak female-\\ strong\\ male trope\end{tabular} &   &   &   &   & 2 & 1 \\ \hline
\end{tabular}%
}
\caption{Pilot 1 results}
\label{tab:pilot_1_csis}
\end{table}


\begin{table}[h!]
\resizebox{\columnwidth}{!}{%
\centering
\begin{tabular}{|l|c|l|c|}
\hline
\large
\textbf{Review Text} & \textbf{Evaluator} & \textbf{Span} & \textbf{Familiarity} \\ \hline
\multirow{10}{*}{\begin{tabular}[c]{@{}l@{}} I always look for the range of \\landscape and plant descriptions\\ in books. Gargash brought the \\realities of inland and coastal \\climate differences to life~- \\especially as related to seasonal\\ changes and shelter~-living\\ in agricultural shelter inland\\ and in urban shelter on the coast.\\ Do you want air conditioning \\or not? Is there such a thing as \\green air conditioning? Is there \\such a thing as sustainable air \\conditioning? Good questions, \\no? \\ \end{tabular}} 
& 1 & {\begin{tabular}[c]{@{}l@{}} living in agri~-\\cultural shelter\\ inland \end{tabular}} & Unfamiliar \\ 
                     & 1 & {\begin{tabular}[c]{@{}l@{}}green air \\ conditioning \end{tabular}} & Unfamiliar \\ 
                     & 1 & {\begin{tabular}[c]{@{}l@{}} sustainable air \\conditioning \end{tabular}} & Unfamiliar \\ \cline{2-4} 
                     & 2 & {\begin{tabular}[c]{@{}l@{}} sustainable air \\conditioning \end{tabular}} & Unfamiliar \\ 
                     & 2 & {\begin{tabular}[c]{@{}l@{}} agricultural \\ shelter \end{tabular}} & Ambiguous \\ 
                     & 2 & urban shelter & Ambiguous \\ \cline{2-4} 
                     & 3 & {\begin{tabular}[c]{@{}l@{}} I always ... shelter \\ on the coast. \end{tabular}} & Unfamiliar \\ 
                     & 3 & {\begin{tabular}[c]{@{}l@{}} Do you ... Good \\questions, no? \end{tabular}} & Unfamiliar \\ \hline
\end{tabular}
}
\caption{Pilot 2 identified spans}
\label{tab:pilot-2-csis}
\end{table}

\begin{table*}[t!]
\centering
\resizebox{\textwidth}{!}{%
\begin{tabular}{|l|l|l|}
\hline
\multicolumn{1}{|c|}{\textbf{Id}} &
  \multicolumn{1}{c|}{\textbf{Span Type}} &
  \multicolumn{1}{c|}{\textbf{Spans}} \\ \hline
0 &
  Cultural &
  \{'Muir woods' 'John Muir?', 'John Muir? Sure, Muir woods,'\} \\
1 &
  Cultural &
  \{'logical meaning ofost', 'ofost'\} \\
2 &
  Non-Cultural &
  \{'"the oath of vayuputras"', 'the title of the book is "oath" of vayuputras! \textbackslash{}n what oath?'\} \\
3 &
  Cultural &
  \{'"People\textbackslash{}'s Republic of China".', 'former state head   of "People\textbackslash{}'s Republic of China".'\} \\

4 &
  Cultural &
  \{'Dalrymple is an old-fashioned curmudgeon',
  'curmudgeon'\} \\ 
5 &
  Cultural &
   \{"'Infinite Jest'.", 'Penguin Ink edition', "attempting   'Infinite Jest'."\} \\
6 &
  Non-Cultural &
  \{'abridged', "I should've went for the abridged version.",   'abridged version.'\} \\
7 &
  Cultural &
   \{'Nature and Selected Essays by Ralph Waldo Emerson'\}\\
8 &
  Cultural &
  \{'The female character is such a classic Heinlein female.',  'classic Heinlein female.', 'a classic Heinlein female.'\} \\
9 &
  Non-Cultural &
  \{'amphibians,', 'ocean acidification, invasive species, ecosystem   fragmentation, poaching,', 'ecosystem fragmentation, poaching'\} \\
10 &
  Cultural &
  \{'Fisssssss...ooouiuu.....!'\} \\
11 &
  Cultural &
   \{'DFW intro,', 'DFW'\} \\
12 & Cultural &  \{"Probably the 'easiest' thing of DFW's that I've read,"\} \\
13 &
  Non-Cultural &
  \{'I myself is a topper from engineering institute'\}\\
14 &
  Non-Cultural &
   \{"militaristic vision",  "sentimentality about War Being Bad."\} \\ \hline
\end{tabular}%
}
\caption{Span texts of the top 15 span clusters identified by the participants across all countries}
\label{tab:top_15_spans}
\end{table*}


\begin{table*}[t!]
\centering
\resizebox{\textwidth}{!}{%
\begin{tabular}{|l|}
\hline
  \multicolumn{1}{|c|}{\textbf{Spans}} \\ \hline
  'tennis lingo', 'western diseases', 'Imperialism Studies', 'poaching', 'anthropologic', 'topper', 'engineering \\ institute', 'bogged down', 'Iraq-centric', 'courtroom dramas' \\ \hline
\end{tabular}
}
\caption{Spans which humans identify as difficult to understand (non-cultural) whereas LLM identifies it as cultural.}
\label{tab:human_noncultural_llm_cultural_filtered}
\end{table*}

\begin{table*}[h!]
\centering
\resizebox{\textwidth}{!}{%
\begin{tabular}{|c|c|l|c|}
\hline
\textbf{\begin{tabular}[c]{@{}c@{}}Participant\\ Location\end{tabular}} &
  \textbf{\begin{tabular}[c]{@{}c@{}}Book\\ Country\end{tabular}} &
  \multicolumn{1}{c|}{\textbf{Span}} &
  \textbf{Span Type} \\ \hline
\multirow{8}{*}{India} &
  India &
  \{'"People\textbackslash{}'s Republic of China".', 'former state head of   "People\textbackslash{}'s Republic of China".''\} &
  Cultural \\
                        & USA   & \{"I understand that Al Gore's new movie ends on a slightly more   upbeat tone,", "Al Gore's new movie"\}                 & Cultural     \\
                        & USA   & \{"Hemingway's Lost Generation...and", "Hemingway's Lost   Generation...and a similar lack of purpose."\}                 & Cultural     \\
                        & USA   & \{'Drum Skin: Fragments on R.A. Harris:', 'Drum Skin: Fragments on R.A.   Harris: "All Art is Junk."'\}                   & Cultural     \\
                        & India & \{'curmudgeon', 'old-fashioned curmudgeon', 'Dalrymple is an old-fashioned   curmudgeon'\}                                & Cultural     \\
                        & USA   & \{"'Infinite Jest'.", 'Penguin Ink edition', "attempting   'Infinite Jest'."\}                                            & Cultural     \\
                        & USA   & \{"her cockatiel's fame", "cockatiel's fame"\}                                                                            & Non-Cultural \\
                        & India & \{'logical meaning ofost', 'logical meaning ofost of the question that   comes to our mind about our mythology.'\}        & Non-Cultural \\ \hline
\multirow{9}{*}{Mexico} & India & \{'Fisssssss...ooouiuu.....!'\}                                                                                           & Cultural     \\
                        & India & \{'eye of a chauffeur.', 'chauffeur.'\}                                                                                   & Cultural     \\
 &
  India &
  \{'"the oath of vayuputras"',    '"oath"', 'the title of the book is "oath" of   vayuputras! \textbackslash{}n what oath?', 'oath?'\} &
  Cultural \\
                        & USA   & \{'DFW intro,', 'DFW'\}                                                                                                   & Cultural     \\
                        & USA   & \{'amphibians,', 'ocean acidification, invasive species, ecosystem   fragmentation, poaching,'\}                          & Non-Cultural \\
                        & USA   & \{'Muir woods,', 'John Muir?','John Muir? Sure, Muir woods,'\}                                                            & Cultural     \\
                        & USA   & \{"'easiest' thing of DFW's that I've read,", "DFW's",   "Probably the 'easiest' thing of DFW's that I've read,"\}        & Cultural     \\
                        & USA   & \{'abridged', "I should've went for the abridged version.",   'abridged version.'\}                                       & Non-Cultural \\
                        & India & \{'I myself is a topper from engineering institute',  'I myself is a topper',\}                                           & Non-Cultural \\ \hline
\multirow{9}{*}{USA} &
  India &
  \{'"the oath of vayuputras"',    '"oath"', 'the title of the book is "oath" of   vayuputras! \textbackslash{}n what oath?', 'oath?'\} &
  Cultural \\
                        & USA   & \{'Nature and Selected Essays by Ralph Waldo Emerson'\}                                                                   & Cultural     \\
                        & India & \{'Medieval Islamic culture,'\}                                                                                           & Cultural     \\
                        & USA   & \{ 'sentimentality about War Being Bad.', ''militaristic vision'\}                                                        & Non-Cultural \\
                        & USA   & \{'Muir woods,', 'John Muir?','John Muir? Sure, Muir woods,'\}                                                            & Cultural     \\
                        & India & \{'especially the parts about Sri Lanka, Pakistan \& Reunion.', 'Sri   Lanka, Pakistan \& Reunion.'\}                     & Cultural     \\
                        & USA   & \{'The female character is such a classic Heinlein female.', 'classic   Heinlein female.', 'a classic Heinlein female.'\} & Cultural     \\
                        & India & \{'I started reading it long after watching the movie 3 idiots', 'the   movie 3 idiots', '3 idiots'\}                     & Cultural     \\
                        & India & \{'a topper from engineering institute', 'topper from engineering',   'topper from engineering institute'\}               & Non-Cultural \\ \hline
\end{tabular}%
}
\caption{Top difficult to understand spans by each participant country}
\label{tab:top_csi_countrywise}
\end{table*}
\onecolumn
\section{Taxonomy Generation and Span Categorization}
\label{appendix:taxonomy_gen_span_categ}
\subsection{Initial Taxonomy Generation Template}
\begin{tcolorbox}[title={Prompt}, width=\textwidth, colback=white, colframe=gray, arc=0pt, outer arc=5pt, boxrule=0.5pt, leftrule=2pt, rightrule=2pt, right=0pt, left=0pt, top=0pt, bottom=0pt, toprule=0pt, bottomrule=2pt]
\label{fig:init_taxonomy}
\small
\#\#\# CONTEXT \\
Annotators were asked to mark spans of text that they found unfamiliar. Below is a list of categories and subcategories. I will provide you with a list of spans, and for each span, please specify the corresponding category and subcategory under which it falls. \\
\{current taxonomy generated by the model\} \\
\#\#\# FULL TEXT \\
\{source text on which spans were marked\} \\
\#\#\# LIST OF SPANS \\
\{list of spans marked by the annotator\} \\
\\
\#\#\# OUTPUT FORMAT \\
\texttt{[[Span 1, Category; Subcategories], ..., [Span X, Category; Subcategories]]} 
\end{tcolorbox}

\subsection{Final Span Categorization And Clustering Template}
\begin{tcolorbox}[title={Prompt}, width=\textwidth, colback=white, colframe=gray, arc=0pt, outer arc=5pt, boxrule=0.5pt, leftrule=2pt, rightrule=2pt, right=0pt, left=0pt, top=0pt, bottom=0pt, toprule=0pt, bottomrule=2pt]
\label{fig:final_span_categorization}
\small
Please categorize the following text spans based on the provided taxonomy and cluster similar spans. \\
Annotators from different demographics marked these spans as unfamiliar or non-understandable from a review text. \\
Please carefully classify each span and provide a brief explanation for your classification. \\
Also, please cluster similar spans. \\
\\ 
\textbf{Taxonomy:} \\
1. Cultural, Linguistic: Differences related to language use, idioms, dialects, or phrasing specific to a culture. \\
2. Cultural, Non-linguistic: Cultural references such as items, traditions, customs, or social norms unrelated to language. \\
3. Non-Cultural, Linguistic: General language issues such as grammar, vocabulary, or syntax that are not culturally specific. \\
4. Non-Cultural, Non-linguistic: Non-cultural factors like complex concepts, technical jargon, or other knowledge-related gaps. \\
5. Poor Quality Annotations: Vague or Ambiguous Annotation, Superfluous Annotation, Marked majority span. \\
\\
\textbf{Task 1:} For each span from the review text below, select the appropriate category from the above taxonomy and briefly explain your reasoning behind the classification in 20 words. \\
\textbf{Task 2:} Cluster the spans which are similar. \\
\\
\textbf{Review text:} \{text\} \\
\textbf{Spans:} \{spans\} \\
\\
Format your answer as a Python dictionary as follows: \\
\{\{"Task 1": \{<span number>: \{'Type': <taxonomy number>, 'Explanation': <explain classification reason in 20 words>\}\}, "Task 2": [<python list of lists of span numbers>]\}\}
\end{tcolorbox}

\onecolumn
\clearpage
\section{Demographic of Annotators from USA, Mexico, India}
\label{appendix:demographics_usa_mex_ind}
\subsection{Demographic of Annotators from USA}
\begin{figure}[h!]
    \centering
    \includegraphics[width=0.4\columnwidth]{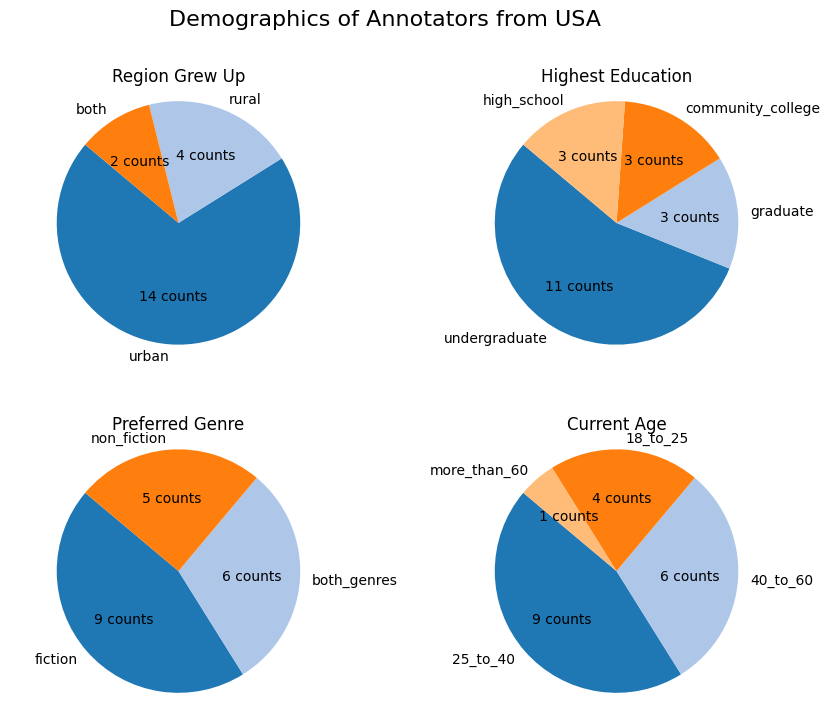}
    \caption{}
    \label{fig:demographics_usa}
\end{figure}

\subsection{Demographic of Annotators from Mexico}
\begin{figure}[h!]
    \centering
    \includegraphics[width=0.4\columnwidth]{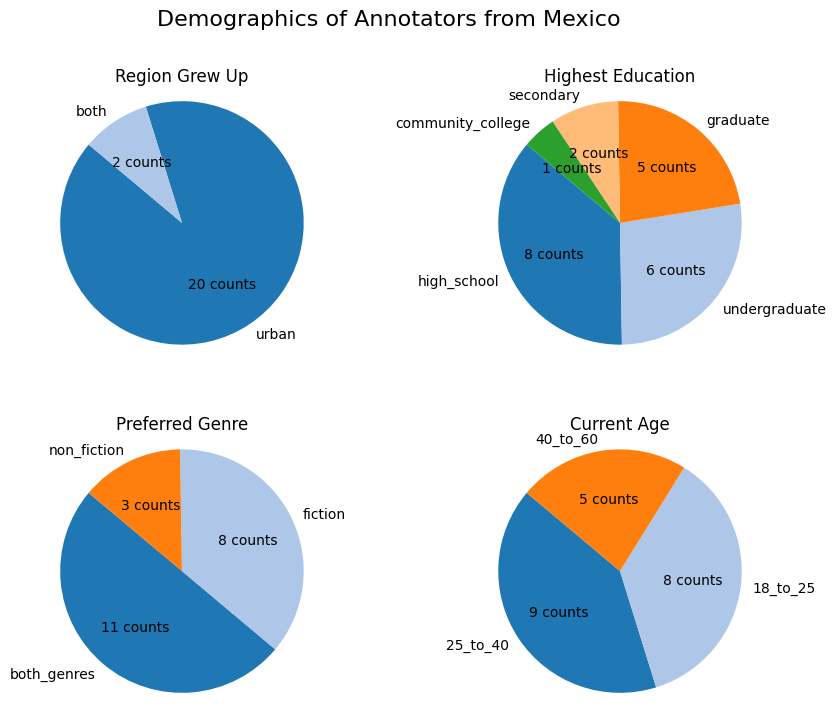}
    \caption{} 
    \label{fig:demographics_mexico}
\end{figure}

\subsection{Demographic of Annotators from India}
\begin{figure}[h!]
    \centering
    \includegraphics[width=0.4\columnwidth]{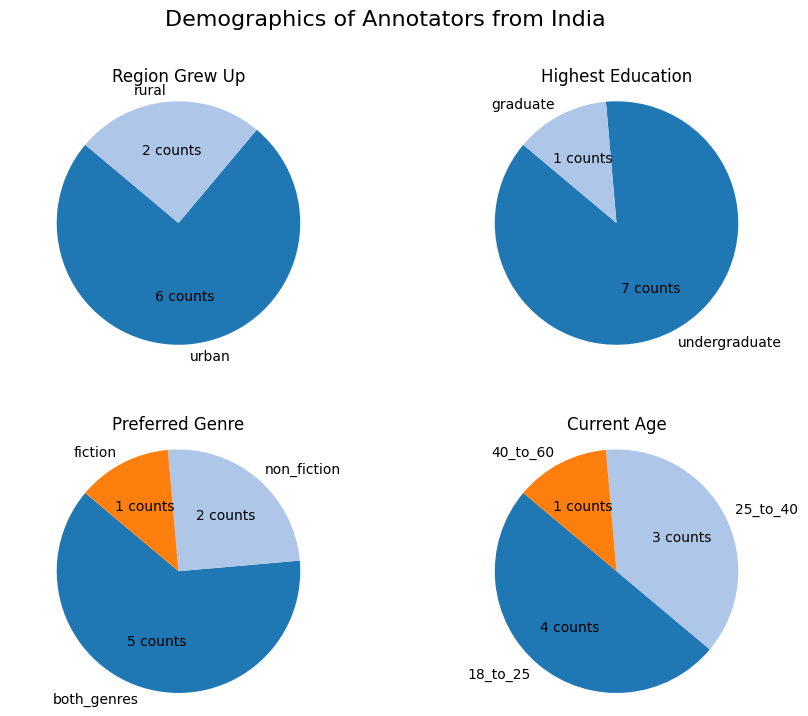}
    \caption{} 
    \label{fig:demographics_india}
\end{figure}

\onecolumn
\clearpage
\section{Label Studio Setup}
\label{appendix:label_studio_setup}
\subsection{Annotation Interface After Signing In}
\label{fig:over_overview}
\begin{figure}[h!]
    \centering
    \includegraphics[width=\columnwidth]{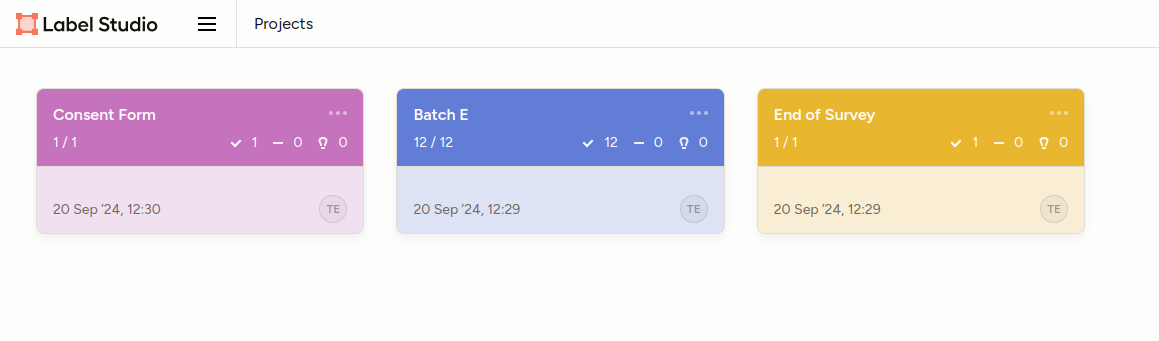}
    \caption{}
\end{figure}

\subsection{Consent, Screener Validation and Demographic Interface}
\label{fig:1_c_sv_di}
\begin{figure}[h!]
    \centering
    \includegraphics[width=\columnwidth]{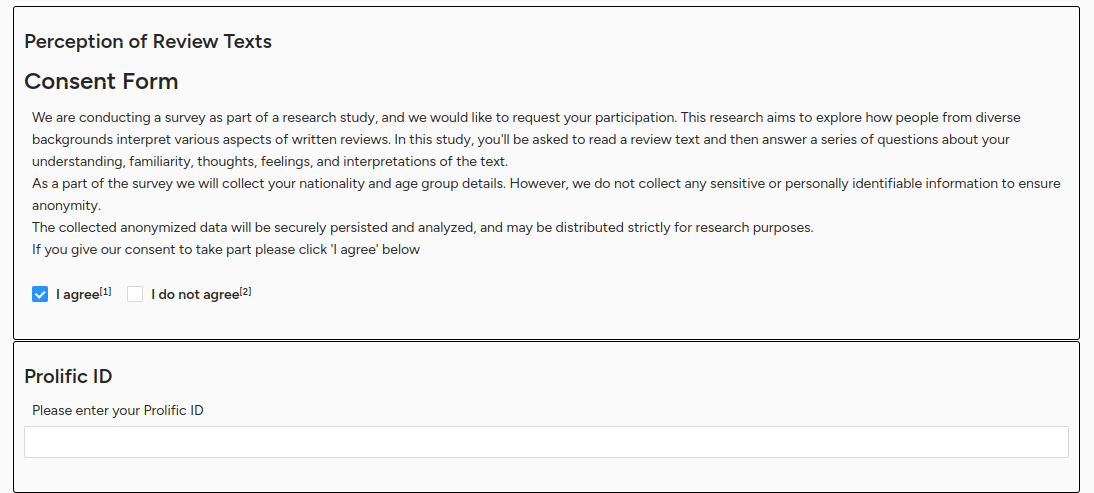}
    \caption{} 
    \label{fig:1_consent}
\end{figure}
\begin{figure}[h!]
    \centering
    \includegraphics[width=\columnwidth]{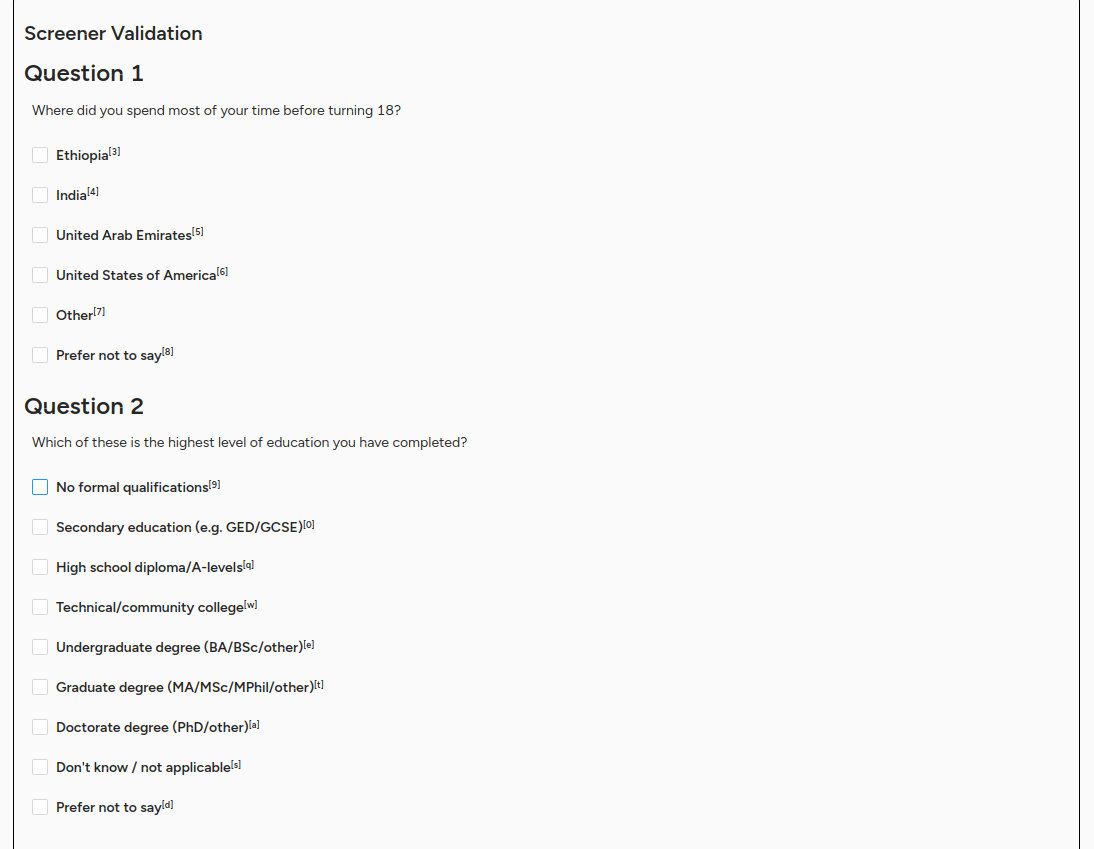}
    \caption{} 
    \label{fig:1_screener_val}
\end{figure}
\begin{figure}[h!]
    \centering
    \includegraphics[width=\columnwidth]{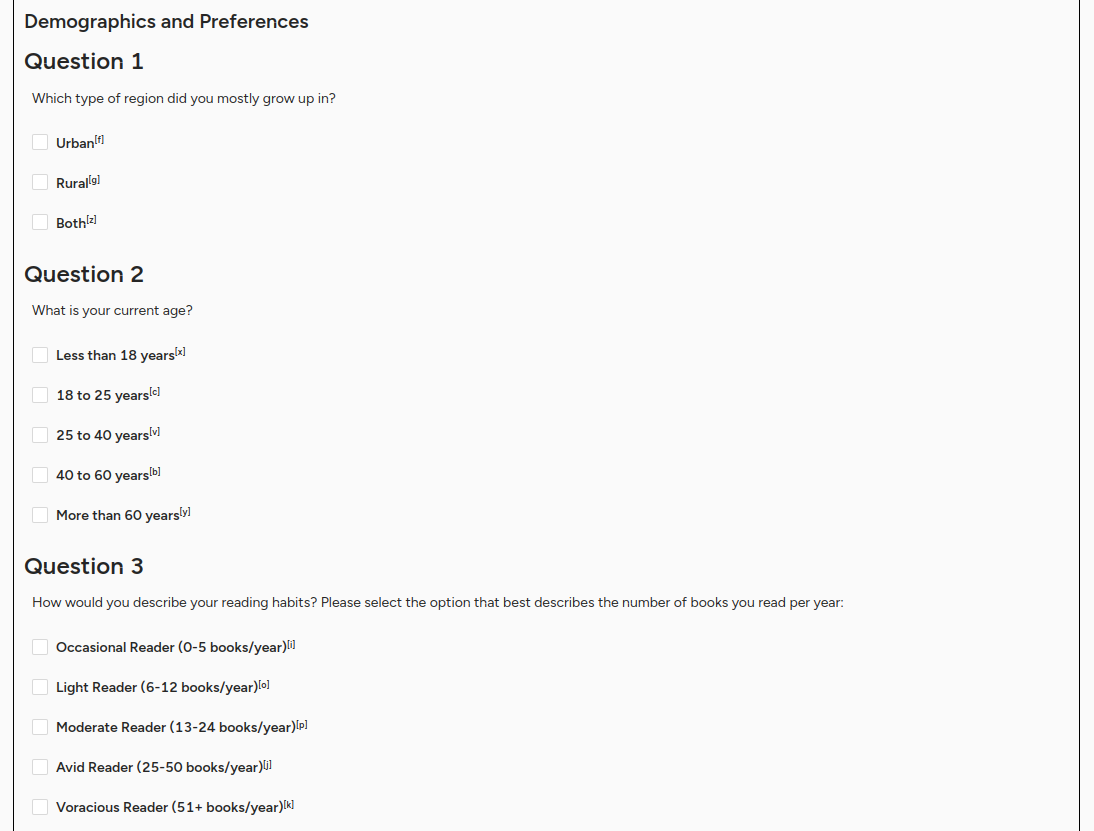}
    \caption{} 
    \label{fig:1_demographics}
\end{figure}

\clearpage
\subsection{Unfamiliarity Span Marking and Other Relevant Questions Interface}
\label{fig:span_mark_other_ques}
\begin{figure}[h!]
    \centering
    \includegraphics[width=\columnwidth]{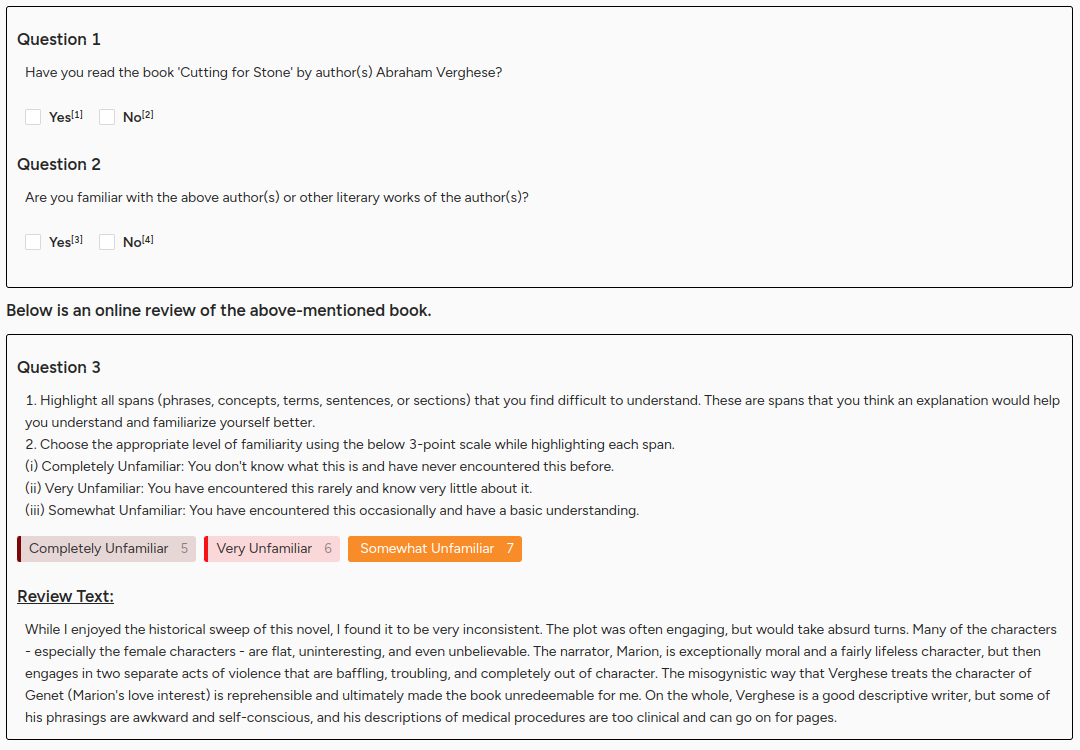}
    \caption{}
\end{figure}
\begin{figure}[h!]
    \centering
    \includegraphics[width=\columnwidth]{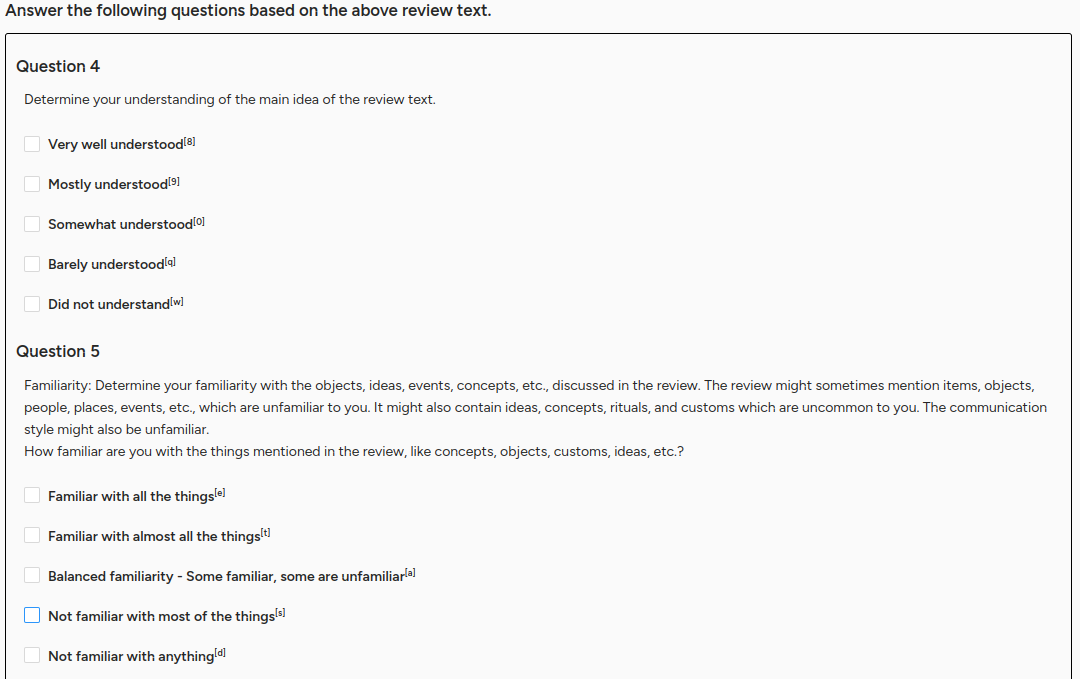}
    \caption{}
\end{figure}
\begin{figure}[h!]
    \centering
    \includegraphics[width=\columnwidth]{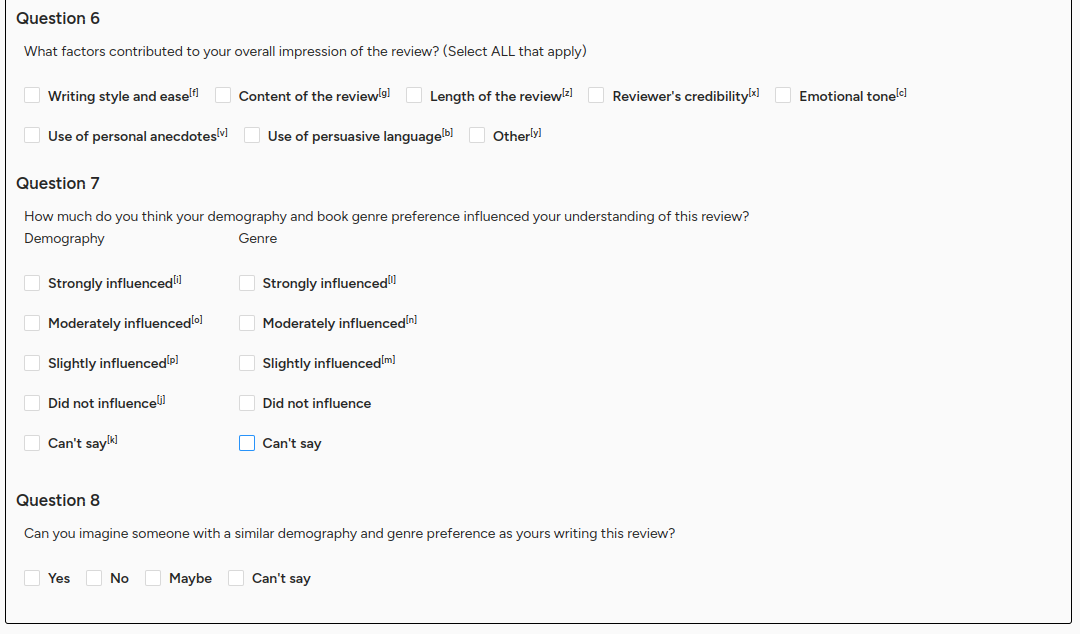}
    \caption{}
\end{figure}

\clearpage
\subsection{End Of Survey Interface}
\label{fig:eos_interface}
\begin{figure}[h!]
    \centering
    \includegraphics[width=\columnwidth]{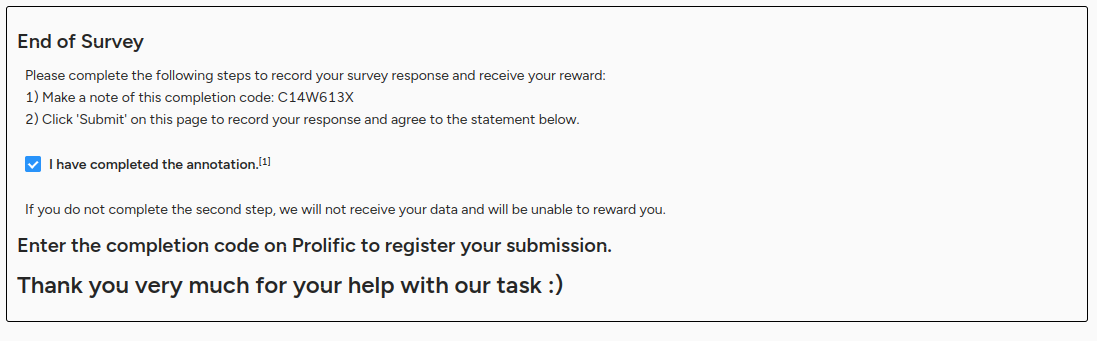}
    \caption{}
\end{figure}


\twocolumn

\section{Additional Analysis}
As depicted in Table \ref{tab:participant-book-review-stats}, participants from India analyzed 26.1, Mexico reviewed 10.5, and the USA reviewed 11.3 reviews on average. Although Indians found more unfamiliar spans (57\%) compared to Mexicans and Americans (45\% each), Americans had the highest proportion of culturally unfamiliar spans (73\%), followed by Mexicans (67\%) and Indians (64\%). Overall, participants from all cultures encountered unfamiliar spans in 50\% of their reviews, and more than two-thirds of them were cultural. This gap in understanding motivates a need for tools that enable cross-cultural communication and provide a headspace for such tools to operate.

Underlined in Table \ref{tab:participant-book-review-stats}, for Indians, the highest proportion of unfamiliar spans (70\%) were for books originating from the USA. Mexicans found the highest proportion of unfamiliar spans, almost equally, from books in India and the USA (approximately 55\%). Americans found unfamiliar spans in 55\% of the Indian book reviews. For participants from all countries, most of the unfamiliarity was for reviews of books from different cultures, which further motivates the need for cross-cultural communication tools.
\begin{table}[h]
\centering
\resizebox{\columnwidth}{!}{%
\begin{tabular}{|l|l|c|c|c|}
\hline
\multicolumn{1}{|c|}{\textbf{\begin{tabular}[c]{@{}c@{}}Participant\\ Country\end{tabular}}} &
  \multicolumn{1}{c|}{\textbf{\begin{tabular}[c]{@{}c@{}}Review\\ Country\end{tabular}}} &
  \textbf{\# Reviews} &
  \textbf{\# Difficult} &
  \textbf{\# Cultural} \\ \hline
\multirow{4}{*}{India}  & Ethiopia \& UAE & 8.00           & 3.29 (41\%)           & 1.86 (57\%)          \\
                        & India           & 7.75           & 4.38 (57\%)           & 2.63 (60\%)          \\
                        & USA             & 10.38          & 7.25 (\underline{70\%})           & 5.00 (69\%)          \\ \cline{2-5} 
                        & \textbf{Total}  & \textbf{26.13} & \textbf{14.92 (57\%)} & \textbf{9.49 (64\%)} \\ \hline
\multirow{4}{*}{Mexico} & Ethiopia \& UAE & 3.28           & 0.72 (22\%)           & 0.39 (54\%)          \\
                        & India           & 3.25           & 1.85 (\underline{57\%})           & 1.25 (68\%)          \\
                        & USA             & 4.00           & 2.18 (\underline{55\%})           & 1.55 (71\%)          \\ \cline{2-5} 
                        & \textbf{Total}  & \textbf{10.53} & \textbf{4.75 (45\%)}  & \textbf{3.19 (67\%)} \\ \hline
\multirow{4}{*}{USA}    & Ethiopia \& UAE & 3.11           & 1.00 (32\%)           & 0.68 (68\%)          \\
                        & India           & 3.45           & 1.90 (\underline{55\%})           & 1.40 (74\%)          \\
                        & USA             & 4.75           & 2.05 (43\%)           & 1.55 (76\%)          \\ \cline{2-5} 
                        & \textbf{Total}  & \textbf{11.31} & \textbf{4.95 (44\%)}  & \textbf{3.63 (73\%)} \\ \hline
\end{tabular}%
}
\caption{Participant location-wise statistics of unfamiliar spans}
\label{tab:participant-book-review-stats}
\end{table}

\clearpage
\section{Benchmarking LLMs}
\label{appendix:benchmarking_llms}

\subsection{Prompts}
\begin{tcolorbox}[title={Prompt}, width=\textwidth, colback=white, colframe=gray, arc=0pt, outer arc=5pt, boxrule=0.5pt, leftrule=2pt, rightrule=2pt, right=0pt, left=0pt, top=0pt, bottom=0pt, toprule=0pt, bottomrule=2pt]
\label{fig:cultural_mediator_prompt}
\small

AI Rules\\
- Output response in JSON format\\
- Do not output any extra text. \\
- Do not wrap the json codes in JSON or Python markers\\
- JSON keys and values in double-quotes\\
    
You are a cultural mediator who understands all cultures across the world. As a mediator, your job is to identify and translate culturally exotic concepts from texts from an unknown source culture to my culture. I am a well-educated \{genre\} lover who grew up in \{article\_urban\} urban \{country\}, which defines my culture.
I came across a review of the book '\{book\}' by \{author\}, which belongs to the \{book\_genre\} genre. Given my cultural background, perform the following tasks:\\

Task 1: Identify all culture-specific items (CSIs) from the review text that I might find hard to understand due to my cultural background. CSIs are textual spans denoting concepts and items uncommon and not prevalent in my culture, making them difficult to understand.\\

Task 2:  For each CSI, identify its category from one of the following seven categories:\\
    1. Ecology: Geographical features, flora, fauna, weather conditions, etc.\\
    2. Material: Objects, artifacts, and products specific to a culture, such as food, clothing, houses, and towns.\\
    3. Social: Hierarchies, practices, and rituals specific to a culture.\\
    4. Customs: Political, social, legal, religious, and artistic organizations and practices. Customs, activities, procedures, and concepts.\\
    5. Habits: Gestures, non-verbal communication methods, and everyday habits unique to a culture.\\
    6. Linguistic: Terms unique to a specific language or dialect, including metaphors, idioms, proverbs, humor, sarcasm, slang, and colloquialisms.\\
    7. Other: Anything not belonging to the above six categories.\\

Task 3:  For each CSI, identify its familiarity from one of the following four levels:\\
    1. Familiar: Most people from my culture know and relate to the concept as intended.\\
    2. Somewhat familiar: Only some people from my culture know and relate to the concept as intended.\\
    3. Unfamiliar: Most people from my culture do not know or relate to the concept.\\
    4. Ambiguous: Most people from my culture know the concept, but its interpretation is varied or conflicting.\\

Task 4:  For each CSI, identify its impact on the readability and understandability of the main point of the entire review text from one of the following three levels:\\
    1. High: Greatly hinders the readability and comprehension of the review, making it difficult to convey its main points effectively.\\
    2. Medium: It somewhat affects the readability and comprehension of the review, leading to only partial conveyance of its content.\\
    3. Low: The review text's readability and comprehension will remain unaffected.\\

Task 5: Within 50 words, detail your reason for highlighting the span as CSI in Task 1 by correlating it with my background.\\

Task 6: Explain each CSI span within 20 words to make it more understandable to me. Provide facts, examples, equivalences, analogies, etc, if needed.\\

Task 7: Reformulate the entire text to make it more understandable to me. Keep the length similar to the original review text.\\

Format your response as a valid Python dictionary formatted as: \{'spans': [List of Python dictionaries where each dictionary item is formatted as: \{'CSI': $<$task 1: copy the CSI span from text$>$, 'category': $<$task 2: CSI category name$>$, 'familiarity': $<$task 3:  familiarity level name$>$, 'impact': $<$task 4: impact level name$>$, 'reason': $<$task 5: reason within 50 words$>$, 'explanation': $<$task 6: explain the span within 20 words$>$\}], 'reformulation': $<$task 7: reformulate entire review text$>$\}. Respond with \{'spans': 'None'\} if you think I will not find anything difficult to understand.\\

Text: \{review\_text\}

\end{tcolorbox}

\clearpage

\end{document}